\documentclass[10pt, a4paper]{article}

\usepackage{lrec-coling2024}
\usepackage{multirow}

\title{Incorporating Lexical and Syntactic Knowledge for Unsupervised Cross-Lingual Transfer}

\name{Jianyu Zheng$^{\ast}$\thanks{${\ast}$ \rm Equal Contribution}$^{\spadesuit}$, Fengfei Fan$^{\ast}$$^{\diamondsuit}$, Jianquan Li$^{\dagger}$\thanks{${\dagger}$ \rm Jianquan Li is the corresponding author}$^{\diamondsuit}$} 

\address{$^{\spadesuit}$Tsinghua University \  $^{\diamondsuit}$Beijing Ultrapower Software Co.,Ltd. \\
	\texttt{zheng\_jianyu@126.com}, \  \texttt{\{fanfengfei, lijianquan2\}@ultrapower.com.cn}}

\abstract{
Unsupervised cross-lingual transfer involves transferring knowledge between languages without explicit supervision. Although numerous studies have been conducted to improve performance in such tasks by focusing on cross-lingual knowledge, particularly lexical and syntactic knowledge, current approaches are limited as they only incorporate syntactic or lexical information. Since each type of information offers unique advantages and no previous attempts have combined both, we attempt to explore the potential of this approach. In this paper, we present a novel framework called "Lexicon-Syntax Enhanced Multilingual BERT" that combines both lexical and syntactic knowledge. Specifically, we use Multilingual BERT (mBERT) as the base model and employ two techniques to enhance its learning capabilities. The \emph{code-switching} technique is used to implicitly teach the model lexical alignment information, while a \emph{syntactic-based graph attention network} is designed to help the model encode syntactic structure. To integrate both types of knowledge, we input code-switched sequences into both the syntactic module and the mBERT base model simultaneously. Our extensive experimental results demonstrate this framework can consistently outperform all baselines of zero-shot cross-lingual transfer, with the gains of 1.0$\sim$3.7 points on text classification, named entity recognition (ner), and semantic parsing tasks.
 \\ \newline \Keywords{cross-lingual transfer, lexicon, syntax, code-switching, graph attention network} }

\begin{document}

\maketitleabstract

\section{Introduction}
Unsupervised cross-lingual transfer refers to the process of leveraging knowledge from one language, and applying it to another language without explicit supervision \cite{conneau2019unsupervised}. Due to the free requirement of the labeled data in target language, it is highly preferred for low-resource scenarios. Recently, unsupervised cross-lingual transfer has been widely applied in various natural language processing (NLP) tasks, such as part-of-speech (POS) tagging \cite{kim2017cross,de2022make}, named entity recognition (NER) \cite{fetahu2022dynamic,xie2018neural}, machine reading comprehension \cite{hsu2019zero,chen2022good}, and question answering (QA) \cite{nooralahzadeh2023improving,asai2021xor}.

The success of unsupervised cross-lingual transfer can be attributed to its ability to exploit connections across languages, which are reflected in various linguistic aspects such as lexicon, semantics, and syntactic structures. Consequently, many studies have sought to enhance models by encouraging them to learn these cross-lingual commonalities. For instance, in the lexical domain, \citet{qin2021cosda} utilize bilingual dictionaries to randomly replace certain words with their translations in other languages, thereby encouraging models to implicitly align representations between the source language and multiple target languages.
In the area of syntax, several works have developed novel neural architectures to guide models in encoding the structural features of languages. \citet{ahmad2021syntax}, for example, proposes a graph neural network (GNN) to encode the structural representation of input text and fine-tune the GNN along with the multilingual BERT (mBERT) for downstream tasks. Both lexical and syntactic approaches facilitate the alignment of linguistic elements across different languages, thereby enhancing the performance of cross-lingual transfer tasks.

\begin{table*}[htbp]
\centering
\resizebox{\textwidth}{!}{
\begin{tabular}{clcccc}
\hline
\multicolumn{1}{l}{\textbf{Lang}} & \textbf{Premise(P)/Hypothesis(H)} & \multicolumn{1}{l}{\textbf{Label}} & \multicolumn{1}{l}{\textbf{+Lex}} & \multicolumn{1}{l}{\textbf{+Syn}} & \multicolumn{1}{l}{\textbf{Ours}} \\
\hline
fr & \begin{tabular}[c]{@{}l@{}} \textbf{P}:Votre société charitable fournit non seulement de les services sociaux communautaires efficaces à \\ les animaux et les personnes, mais sert également également de fourrière pour la Ville de Nashua.\\  \textbf{H}:La société humaine est le refuge pour animaux de Nashua.\end{tabular} & entali  & {\color[HTML]{CB0000} contra}  & {\color[HTML]{CB0000} contra}     & {\color[HTML]{A9D18E} entail}     \\
de & \begin{tabular}[c]{@{}l@{}} \textbf{P}:Ihre humane Gesellschaft erbringt nicht nur effektive gemeinschaftlich-soziale Dienstleistungen \\ für Tiere und ihre Menschen, sondern dient auch als Zwinger der Stadt Nashua.\\ \textbf{H}:Die Humane Society ist Nashuas Tierheim.\end{tabular} & entail  & {\color[HTML]{CB0000} contra}  & {\color[HTML]{CB0000} contra}     & {\color[HTML]{A9D18E} entail}     \\
en & \multicolumn{5}{l}{\begin{tabular}[c]{@{}l@{}} \textbf{P}:Your humane society provides not only effective community social services for animals and their \\ people , but also serves as the pound for the City of Nashua .\\ \textbf{H}:The humane society is Nashua's animal shelter .\end{tabular}}  \\                                         \hline                                                                                                
\end{tabular}
}
\caption{
The parallel sentence pairs in French and German from XNLI\cite{conneau2018xnli}, which are translated from English. Each sentence pair consist of a Premise sentence(P) and a Hypothesis sentence(H). The "Label" column indicates the relationship between each sentence pair, which can be contradiction(contra), entailment(entail) or neutral. "+Lex" and "+Syn" represent the prediction results from the multilingual models infused with lexical and syntactic knowledge, respectively. The "ours" column shows the results of integrating both types of knowledge into the model. Compared to the other two methods, our method can accurately predict the relationship between each sentence pair.
}
\label{table1}
\end{table*}

However, language is a highly intricate system \cite{ellis2009language}, with elements at various levels being interconnected. For example, sentences are composed of phrases, which in turn are composed of words. In cross-lingual transfer, we hypothesize that merely guiding models to focus on a single linguistic aspect is inadequate. Instead, by simultaneously directing models to learn linguistic knowledge across diverse levels, their performance can be further improved.
Table~\ref{table1} presents some example sentences extracted from the XNLI dataset \cite{conneau2018xnli}. These parallel sentence pairs demonstrate that the multilingual model makes incorrect predictions for sentence pairs in the target languages (French and German) when only one aspect of linguistic knowledge, such as lexical or syntactic knowledge, is incorporated. However, when both types of knowledge are integrated into the model, the correct prediction is obtained. Despite this, most previous studies have focused on either syntactic or lexical information alone, without considering the integration of both types of information.

In this work, we aim to enhance unsupervised cross-lingual transfer by integrating knowledge from different linguistic levels. To achieve this, we propose a framework called "Lexicon-Syntax Enhanced Multilingual BERT" ("LS-mBERT"), based on a pre-trained multilingual BERT model. Specifically, we first preprocess the input source language sequences to obtain each word's part-of-speech information and dependency relationships between words in each sentence. Then, we replace some words in the sentence with their translations from other languages while preserving the established dependency relationships. Furthermore, we employ a graph attention network\cite{velivckovic2017graph} to construct a syntactic module, the output of which is integrated into the attention heads of the multilingual BERT. This integration guides the entire model to focus on syntactic structural relationships. Finally, during the fine-tuning process, we simultaneously train the multilingual BERT and the syntactic module with the pre-processed text. As a result, our framework enables the multilingual BERT to not only implicitly learn knowledge related to lexical alignment but also encode knowledge about syntactic structure.

To validate the effectiveness of our framework, we conduct experiments on various tasks, including text classification, named entity recognition (ner), and semantic parsing. The experimental results show that our framework consistently outperforms all baseline models in zero-shot cross-lingual transfer across these tasks. For instance, our method achieves the improvement of 3.7 points for mTOP dataset. Our framework also demonstrates significant improvements in generalized cross-lingual transfer. Moreover, we examine the impact of important parameters, such as the replacement ratio of source words, and languages for replacement. To facilitate further research explorations, we release our code at \url{https://github.com/Tian14267/LS_mBert}.

\section{Related Work}
Cross-lingual transfer is crucial in the field of natural language processing (NLP) as it enables models trained on one language to be applied to another. To enhance performance in transfer tasks, numerous studies focus on addressing the characteristics of various languages and their relationships.

\subsection{Incorporating Lexical Knowledge for Cross-lingual Transfer}
A group of studies aims to incorporate lexical alignment knowledge into cross-lingual transfer research \cite{zhang2021point, wang2022expanding,qin2021cosda,lai2021saliency}. For example, \citet{zhang2021point} and \citet{wang2022expanding} employ bilingual dictionaries to establish word alignments and subsequently train cross-lingual models by leveraging explicit lexical associations between languages. Other methods \cite{qin2021cosda, lai2021saliency} involve substituting a portion of words in a sentence with their equivalents from different languages, a technique commonly known as "code-switching." By increasing the diversity of input text, these approaches promote implicit alignments of language representations. However, this group of studies mainly offers insights into lexical translation across languages, while neglecting the learning of language-specific structural rules.

\subsection{Incorporating Syntactic Knowledge for Cross-lingual Transfer}
Another research category focuses on integrating syntactic knowledge for cross-lingual transfer \cite{ahmad2021syntax, yu2021cosy, zhang2021benefit, he2019syntax, cignarella2020multilingual, xu2022zero, shi2022substructure, wang2021cross}. Many studies in this group \cite{ahmad2021syntax, wang2021cross} develop graph neural networks to encode syntactic structures, a category to which our work also belongs. Taking inspiration from \citet{ahmad2021syntax}, we adopt a similar architecture, specifically using a graph attention network to encode syntactic knowledge. Other methods \cite{cignarella2020multilingual, xu2022zero} extract sparse syntactic features from text and subsequently incorporate them into the overall model. Although these approaches consider the relationships between language elements, they frequently overlook the alignments across languages, which impedes the effective transfer of linguistic elements and rules between languages.

Consequently, we combine the strengths of these two categories of approaches. First, we replace the input sequence with translated words from other languages, which aids in guiding the entire model to acquire implicit alignment information. Then, we introduce an additional module to assist the model in encoding syntax. 

\begin{figure*}[htbp]
\centering
\includegraphics[width=1\textwidth,height=0.30\textwidth]{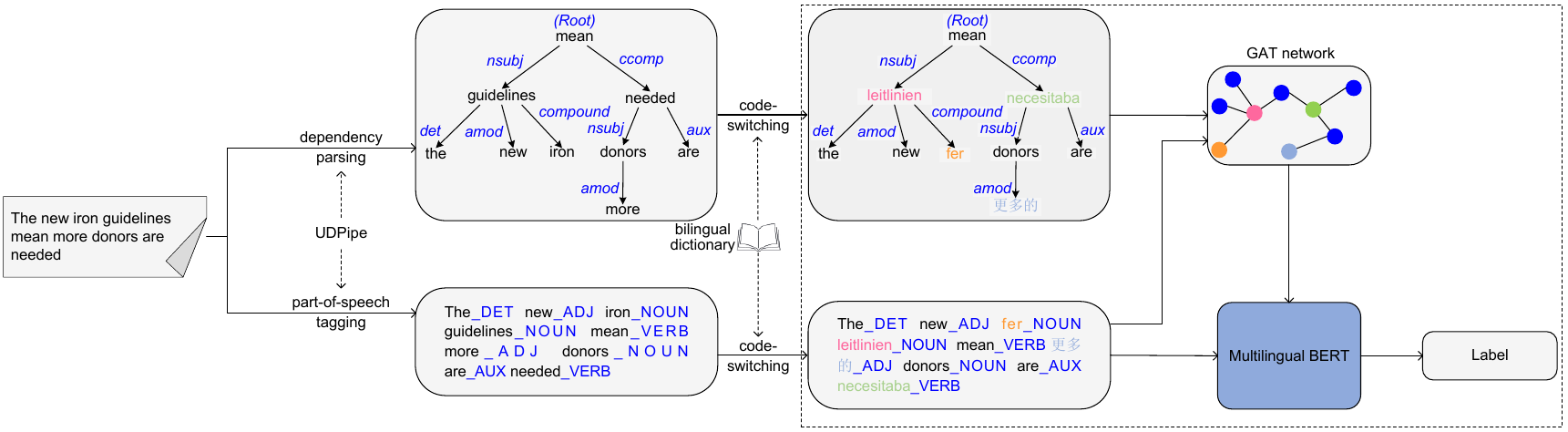} 
\caption{An overview of lexicon-syntax enhanced multilingual BERT ("LS-mBERT"). An example sentence is provided to explain how this framework works. To introduce lexical alignment knowledge, we utilize bilingual dictionaries to randomly replace some words in the sentence with the equivalent words from other languages (pink for German, green for Spanish, light blue for Chinese, and orange for French). Then, an graph attention network (GAT) is developed to encode the syntactic structure of this sentence. The output representation of GAT is sent to the attention heads in multilingual BERT for guiding them to focus on the language-specific structures.}
\label{fig.1}
\end{figure*}

\section{Methodology}
In this section, we provide a detailed introduction to our framework "LS-mBERT", as illustrated in Figure~\ref{fig.1}. Our objective is to enhance the cross-lingual transfer capabilities of multilingual BERT (mBERT) by incorporating both lexical and syntactic knowledge. Given an input sequence, we first pre-process it using a part-of-speech tagger and a universal parser(Section~\ref{sec3.1}). This yields the part-of-speech tag for each word and dependency relationships among words in the sequence. To enable mBERT to implicitly encode word alignment information, we substitute some words with their translations from other languages using a code-switching technology (Section~\ref{sec3.2}). Moreover, to guide mBERT in attending to syntactic relationships, we construct a graph attention network (GAT), introduced in Section~\ref{sec3.3}. The output of the graph attention network is then used as input to the attention heads within BERT, effectively biasing attention information between words. Finally, to integrate both syntactic and lexical knowledge, we pass the code-switched text into both the GAT network and mBERT, which are trained simultaneously (Section~\ref{sec3.4}).

\subsection{Pre-processing Input Sequence}
\label{sec3.1}
The initial step involves pre-processing the input data to obtain prior knowledge for subsequent training. As our framework incorporates syntactic knowledge, we opt for an off-the-shelf parser with high accuracy to process the input text. In this case, we employ the UDPipe toolkit\cite{straka2017tokenizing} to parse the inputs sentences, and Stanza\cite{qi2020stanza} to annotate the part-of-speech information of each word. By utilizing both tools, given a sentence, we can obtain the dependency relationships between words and their part-of-speech information, which are then utilized to provide syntactic knowledge and enhance word representations, respectively.

\subsection{Code-switching for Text (lexical knowledge)}
\label{sec3.2}
As our objective is to improve unsupervised cross-lingual transfer, introducing explicit alignment signals would be inappropriate. Therefore, we employ an implicit strategy to guide the entire model to encode word alignment information. Inspired by the work of \citet{qin2021cosda}, we opt for the code-switching strategy. Specifically, we first randomly select a proportion $\alpha$ of words within each source sentence. Then, for each selected word, we use a high-quality bilingual dictionary to substitute it with a corresponding translation from another target language. This method not only promotes the implicit alignment of representations across diverse languages within our model, but also enhances the model's robustness when processing input text.

\subsection{Graph Attention Network (syntactic knowledge)}
\label{sec3.3}
\begin{figure}[htbp]
\centering
\includegraphics[width=0.49\textwidth,height=0.38\textwidth]{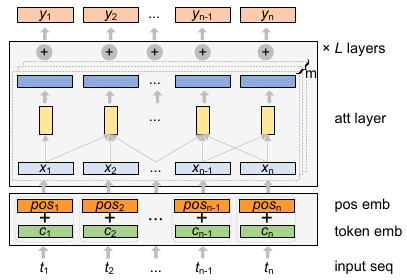} 
\caption{The architecture of graph attention network \cite{ahmad2021syntax,velivckovic2017graph}. Each input token is represented by combining its token embedding and part-of-speech embedding. Each attention head within the graph attention network(GAT) generates a representation for each token embedding by attending to its neighboring tokens in the dependency graph. Next, the resulting representations are concatenated to form the output representation for each token. Finally, we can obtain the representations of the output sequence embeddings from the final layer of GAT.}
\label{fig.2}
\end{figure}

To guide mBERT in acquiring syntactic knowledge better, we construct an external syntactic module by referring to the method introduced by \citet{ahmad2021syntax}. The overview of this module is displayed in Figure~\ref{fig.2}. Given that there are $n$ tokens in the input sequence, we first represent each token by combining its embedding representation with part-of-speech (POS) information. The representation of the $i$-th token can be calculated: $x_i=c_iW_c+pos_iW_{pos},$
where $c_i$ and $pos_i$ represent the token representation and the part-of-speech representation of the $i$-th token, respectively; while $W_c$ and ${W}_{pos}$ denote the token parameter matrix and the part-of-speech parameter matrix.
Then, the encoded sequence $s' = [x_1, x_2, \cdots , x_n]$ is passed into the subsequent syntactic module, which is designed with a graph attention network (GAT) \cite{velivckovic2017graph}. The GAT module comprises a total of $L$ layers, each with $m$ attention heads. These attention heads play a crucial role in generating representations for individual tokens by attending to neighboring tokens in the graph. Each attention in GAT operates as follows: $O = Attention(T, T, V, M)$, wherein $T$ denotes the query and key matrices, and $V$ represents the value matrix.  Besides, $M$ signifies the mask matrix, determining whether a pair of words in the dependency tree can attend each other. Notably, the relationships between words in the attention matrix are modeled based on the distances between words in the dependency tree, rather than the positional information within the word sequence. Subsequently, the resulting representations produced by all attention heads are concatenated to form the output representations for each token. Finally, the output sequence from the final layer can be denoted as $Y = [y_1, y_2, \cdots, y_n]$, where $y_i$ represents the output representation for the $i$-th token.
To maintain the lightweight nature of the architecture, certain elements in GAT have been excluded. Specifically, we do not employ feed-forward sub-layers, residual connections, or positional representations. We found that these modifications do not result in a significant performance gap.

\subsection{Summary of the Framework: Lexicon-syntax Enhanced Multilingual BERT}
\label{sec3.4}
In this subsection, we provide an overview of our "LS-mBERT" framework, as illustrated in Figure~\ref{fig.1}. We first select multilingual BERT (mBERT) as the base model. Then, we process the input sequence using the code-switching strategy in Section~\ref{sec3.2}, resulting in the code-switched sequence $s'$. It is important to note that despite some words in each sentence being replaced with other languages, the original dependency relationships between words are still preserved in $s'$. Next, we feed the code-switched text into both mBERT and the syntactic module (GAT), facilitating the fusion of the two types of knowledge. Furthermore, this step guides the entire model to better align different languages within the high-dimensional vector space during training. After GAT processes the code-switched sequence, the output from the final layer is utilized to bias the attention heads of mBERT. The calculation process can be described as follows: $O=Attention(Q+YW_{l}^{Q},K+YW_{l}^{K},V)$, where $Q$, $K$, and $V$ represent the query, key, and value matrices, respectively; While $W_{l}^{Q}$ and $W_{l}^{K}$ are new parameters to learn for biasing the query and key matrices. 

\section{Experiments} 
\subsection{Experimental Settings}
As above mentioned, we use UDPipe \cite{straka2017tokenizing} and Stanza \cite{qi2020stanza} for parsing sentences and obtaining words' part-of-speech information in all languages, and employ MUSE \cite{lample2018word} as the bilingual dictionary for word substitution. For all tasks, we identify the optimal parameter combinations by searching within the candidate sets. The learning rate is set to 2e-5, utilizing $AdamW$ as the optimizer. The batch size is 64, and the maximum length for input sequences is 128 tokens. For code-switching, we vary the replacement ratio ($\alpha$) from 0.3 to 0.7 with a step of 0.1. For the GAT network, we adopt the identical parameter values as employed in the work of \citet{ahmad2021syntax}. Specifically, we set $L$ to 4 and $k$ to 4.

\subsection{Tasks}

Our framework is evaluated on the following tasks, using English as the source language. Some statistics are summarized in Table~\ref{table2}, along with the detailed descriptions provided below.

\noindent{\bf Text Classification.} Text Classification is a task that assigns predefined categories to open-ended text. In our experiment, we utilize two publicly available dataset: XNLI and PAWS-X. In XNLI \cite{conneau2018xnli}, models need to predict whether a given pair of sentences is entailed, contradicted, or neutral; In PAWS-X \cite{yang2019paws}, models are required to determine whether two given sentences or phrases convey the same meaning. When implementing the two tasks, to establish connections between the dependency trees of the two sentences, we introduce two edges from the $[CLS]$ token to the root nodes. Subsequently, we apply the code-switching technique to randomly replace certain words in the sentence pairs.

\noindent{\bf Named Entity Recognition.} Named Entity Recognition (NER) is a task that involves the automatic identification and categorization of named entities. In our experiment, we employ the Wikiann \cite{pan2017cross} dataset. Wikiann consists of Wikipedia articles annotated with person, location, organization, and other tags in the IOB2 format. Our method is evaluated across 15 languages. To ensure that the models can obtain complete entity information, we exclusively substitute words that do not constitute named entities during the code-switching process.

\noindent{\bf Task-oriented Semantic Parsing.} In this task, the models are required to determine the intent of the utterance and then fill the relevant slots. The dataset for the experiment is mTOP \cite{li2021mtop}, which is an almost parallel corpus, containing 100k examples in total across 6 languages. Our experiments cover 5 languages.

\begin{table*}[htbp]
\centering
\begin{tabular}{lllllll}
\hline
\textbf{Task}    & \textbf{Dataset} & \textbf{|Train|} & \textbf{|Dev|} & \textbf{|Test|} & \textbf{|Lang|} & \textbf{Metric} \\
\hline
Classification& XNLI& 392K& 2.5K& 5K& 13& Accuracy\\
Classification& PAWS-X& 49K& 2K& 2K& 7& Accuracy\\
NER& Wikiann& 20K& 10K& 1-10K& 15& F1\\
Semantic Parsing& mTOP& 15.7K& 2.2K& 2.8-4.4K& 5& Exact Match\\    
\hline
\end{tabular}
\caption{
Evaluation datasets. |Train|, |Dev| and |Test| delegate the numbers of examples in the training, validation and testing sets, respectively. |Lang| is the number of target languages we use in each task.
}
\label{table2}
\end{table*}

\subsection{Baselines}
We choose the following methods as baselines to compare:
\begin{itemize}
\item {\bf mBERT.} We exclusively utilize the multilingual BERT model to perform zero-shot cross-lingual transfer for these tasks. 
\item {\bf mBERT+Syn.} A graph attention network (GAT) is integrated with multilingual BERT, and these two components are jointly trained for all tasks. 
\item {\bf mBERT+Code-switch.} The multilingual BERT model is fine-tuned with the code-switched text across various languages. 
\end{itemize}



\section{Results and analysis}
\subsection{Cross-Lingual Transfer Results}
The main experimental results are displayed in Table~\ref{table3}. Our method consistently demonstrates superior performance across all tasks compared to other baselines. This indicates our method's effectiveness for cross-lingual transfer, achieved through the incorporation of lexical and syntactic knowledge. Especially for the tasks Wikiann and mTOP, our method exhibits a significant improvement, with an increase of 2.2 and 3.7 points, respectively, when compared to the baseline with the best performance.  In addition, since code-switching technique blends words from various language, we calculate the results across the languages excluding English, as shown in the column "AVG/en" in Table~\ref{table3}. We find that the performance gap between our method and each baseline in most tasks becomes wider. This also indicates that our method can more effectively align non-English languages within the same vector space implicitly.  

For each task, we discover most of languages can gain improvement by using our method, as compared to the top-performing baseline. Specifically, 84.6$\%$ (11/13), 100.0$\%$ (7/7), 80.0$\%$ (12/15) and 100.0$\%$ (5/5) languages demonstrate improvement in XNLI, PAWS-X, Wikiann and mTOP respectively. Furthermore, our method also provides improvement for non-alphabetic languages in many tasks, such as Chinese, Japan and Korean. This reflects that our method can be effectively generalized into various target languages, even in cases where significant differences exist between the source and target languages.

\begin{table*}[!htbp]
\centering
\resizebox{\textwidth}{!}{
\begin{tabular}{llccccccccccccccccccc}
\hline
\multicolumn{1}{c}{\textbf{Tasks}} & \multicolumn{1}{c}{\textbf{Methods}} & \multicolumn{1}{c}{\textbf{en}} & \multicolumn{1}{c}{\textbf{ar}} & \multicolumn{1}{c}{\textbf{bg}} & \multicolumn{1}{c}{\textbf{de}} & \multicolumn{1}{c}{\textbf{el}} & \multicolumn{1}{c}{\textbf{es}} & \multicolumn{1}{c}{\textbf{fr}} & \multicolumn{1}{c}{\textbf{hi}} & \multicolumn{1}{c}{\textbf{ru}} & \multicolumn{1}{c}{\textbf{tr}} & \multicolumn{1}{c}{\textbf{ur}} & \multicolumn{1}{c}{\textbf{vi}} & \multicolumn{1}{c}{\textbf{zh}} & \multicolumn{1}{c}{\textbf{ko}} & \multicolumn{1}{c}{\textbf{nl}} & \multicolumn{1}{c}{\textbf{pt}} & \multicolumn{1}{c}{\textbf{ja}} & \multicolumn{1}{c}{\textbf{AVG / en}} & \multicolumn{1}{c}{\textbf{AVG}} \\
\hline
\multirow{4}{*}{XNLI \cite{conneau2018xnli}}& mBERT& 80.8& 64.3& 68.0& 70.0& 65.3& 73.5& 73.4& 58.9& 67.8& 60.9& 57.2& 69.3& 67.8& -& -& -& -& 66.4& 67.5\\
& mBERT+Syn& \textbf{81.6}& 65.4& 69.3& 70.7& 66.5& 74.1& 73.2& 60.5& 68.8& \textbf{62.4}& 58.7& 69.9& 69.3& -& -& -& -& 67.4 & 68.5\\
& mBERT+code-switch& 80.9& 64.2& 70.0& 71.5& 67.1& 73.7& 73.2& 61.6& 68.9& 58.6& 57.8& 69.9& 70.0& -& -& -& -& 67.2& 68.3\\
& our method& 81.3& \textbf{65.8}& \textbf{71.3}& \textbf{71.8}& \textbf{68.3} & \textbf{75.2}& \textbf{74.2}& \textbf{62.8}& \textbf{70.7}& 61.1& \textbf{58.8}& \textbf{71.8}& \textbf{70.8}& -& -& -& -& \textbf{68.6}& \textbf{69.5}\\
\hline
\multirow{4}{*}{PAWS-X \cite{yang2019paws}}& mBERT& 94.0& -& -& 85.7& -& 87.4& 87.0& -& -& -& -& -& 77.0& 69.6& -& -& 73.0& 80.2& 81.7\\
& mBERT+Syn& 93.7& -& -& 86.2& -& 89.5& 88.7& -& -& -& -& -& 78.8& 75.5& -& -& 75.9& 82.7& 83.9\\
& mBERT+code-switch & 92.4& -& -& 85.9& -& 87.9& 88.3& -& -& -& -& -& 80.2& 78.0& -& -& 78.0& 83.4& 84.3\\
& our method& \textbf{93.8}& -& -& \textbf{87.2}& -& \textbf{89.6}& \textbf{89.4}& -& -& -& -& -& \textbf{81.8}& \textbf{79.0}& -& -& \textbf{80.0}& \textbf{84.6}& \textbf{85.6}\\
\hline
\multirow{4}{*}{Wikiann\cite{pan2017cross}}& mBERT& 83.7& 36.1& 76.0& 75.2& 68.0& 75.8& 79.0& 65.0& 63.9& 69.1& 38.7& 71.0& -& 58.9& 81.3& 79.0& -& 66.9& 68.1\\
& mBERT+Syn& 84.1& 34.6& 76.9& 75.4& 68.2& \textbf{76.0}& 79.1& 64.0& \textbf{64.2}& 68.7& 38.0& \textbf{73.1}& -& 58.0& 81.7& 79.5& -& 67.0& 68.1\\
& mBERT+code-switch& 82.4& 39.2& 77.1& 75.2& 68.2& 71.0& 78.0& 66.1& \textbf{64.2}& 72.4& 41.3& 69.2&& 59.9& 81.3& 78.9& -& 67.3& 68.3\\
& our method&\textbf{84.5}&\textbf{41.4}&\textbf{78.9}&\textbf{77.3}&\textbf{70.2}&75.3&\textbf{80.3}&\textbf{67.6}&63.9&\textbf{73.1}& \textbf{46.8}&72.6 & - & \textbf{62.2}& \textbf{81.8} & \textbf{80.8}& -& \textbf{69.4} & \textbf{70.5} \\
\hline
\multirow{4}{*}{mTOP\cite{li2021mtop}}& mBERT& 81.0& -& -& 28.1& -& 40.2& 38.8& 9.8& -& -& -& -& -& -& -& -& -& 29.2& 39.6\\
& mBERT+Syn& 81.3& -& -& 30.0& -& 43.0& 41.2& 11.5& -& -& -& -& -& -& -& -& -& 31.4& 41.4\\
& mBERT+code-switch& 82.3& -& -& 40.3& -& 47.5& 48.2& 16.0& -& -& -& -& -& -& -& -& -& 38.0& 46.8\\
& our method& \textbf{83.5}& -& -& \textbf{44.5}& -& \textbf{54.2}& \textbf{51.7}& \textbf{18.8}& -& -& -& -& -& -& -& -& -& \textbf{47.3}& \textbf{50.5}\\        
\hline
\end{tabular}
}
\caption{
The experimental results on four tasks. The best results in each task are highlighted in bold. The baselines include "mBERT", "mBERT+Syn" and "mBERT+codeswitch". They delegate "only using mBERT", "using mBERT with a syntactic module (GAT)" and "mBERT with the code-switching technique" for cross-lingual transfer. The results of "mBERT" is from \citet{hu2020xtreme}. For "mBERT+Syn" and "mBERT+code-switch", we adopt open-source code of the work of 
\citet{ahmad2021syntax} and  \citet{qin2021cosda} to reproduce these experiments, and report the results. The evaluation metrics are F1 value for the NER task, Accuracy for classification tasks, and Exact Match for semantic parsing. The "AVG" column means the average performance across all language for each method, while the "AVG /en" indicates the average performance on the languages excluding English.
}
\label{table3}
\end{table*}

\subsection{Generalized Cross-Lingual Transfer Results}
\begin{figure*}[htbp]
\centering
\includegraphics[width=0.93\textwidth,height=0.38\textwidth]{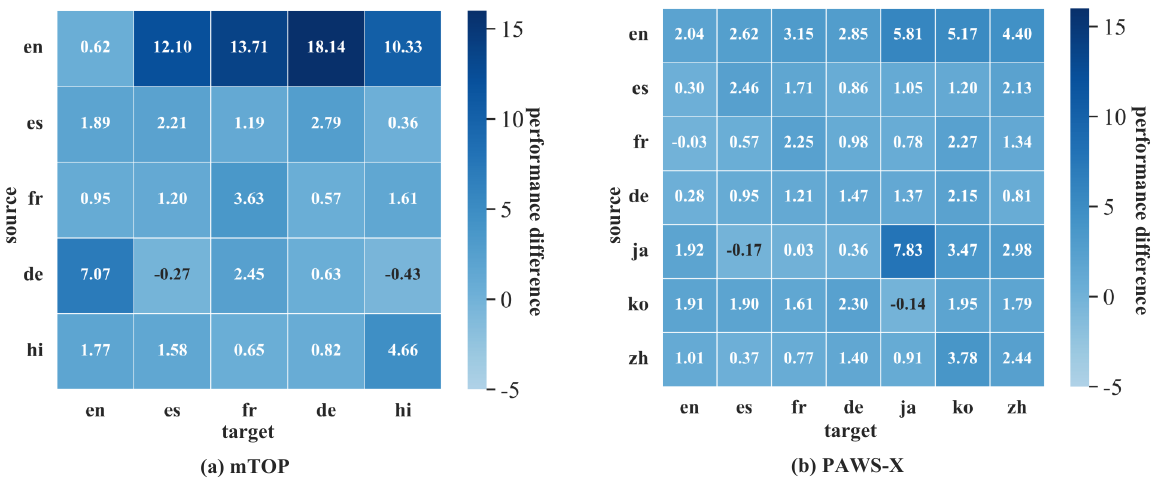} 
\caption{Results for generalized zero-shot cross-lingual transfer on mTOP and PAWS-X. We report the performance differences between our method and "mBERT" baseline across all languages.}
\label{fig.3}
\end{figure*}

In practical scenarios, cross-lingual transfer could involve any language pair. For example, in a cross-lingual question-answering (QA) task, the context passage may be in German, while the multilingual model is required to answer the question in French. Considering on this, we conduct zero-shot cross-lingual transfer experiments within a generalized setting. Since PAWS-X and mTOP are completely parallel, we evaluate the performance of our method and "mBERT" baseline on generalized cross-lingual transfer tasks using the two dataset. The experimental results are illustrated in Figure~\ref{fig.3}.

For both classification and semantic parsing benchmarks, we have observed improvements among most language pairs. This reflects that our method is very effective for generalized cross-lingual transfer. Furthermore, when English is included in the language pair, there is a substantial enhancement in performance. Specifically, when English serves as the source language, the average performance of target languages is increased over 10$\%$ and 3$\%$ in mTOP and PAWS-X dataset, respectively. This reflects the effectiveness of the code-switching in aligning other languages with English. For the PAWS-X dataset, we find that some non-Indo-European languages such as Japanese, Korean, and Chinese can achieve improvements, even when the source languages belong to the Indo-European language family, including English, Spanish, French, and German. It reflects that syntactic knowledge can effectively narrow the gap of language structures for this task, especially for the language pairs without close linguistic relationships.

\section{Analysis and Discussion}
\subsection{Impact on Languages}
\begin{figure*}[htbp]
\centering
\includegraphics[width=0.76\textwidth,height=0.30\textwidth]{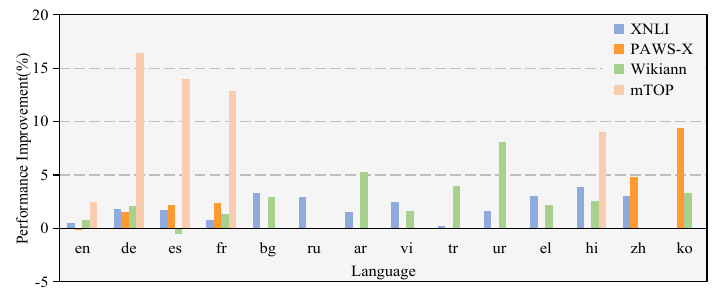} 
\caption{Performance improvements for XNLI, PAWS-X, Wikiann, and mTOP across languages. The languages in x-axis are grouped by language families: IE.Germanic (en, de), IE.Romance (es, fr), IE.Slavic (bg, ru), Afro-asiatic (ar), Austro-asiatic (vi), Altaic (tr, ur), IE.Greek (el), IE.Indic (hi), Sino-tibetan (zh), Korean (ko).}
\label{fig.4}
\end{figure*}

We investigate whether our method can improve the performance of specific languages or language groups. As shown in Figure~\ref{fig.4}, we display the performance improvement of our method by comparing the "mBERT" baseline. We find that almost languages can obtain benefits from our method. Particularly, when the target language, such as German, Spanish and French, belongs to the Indo-European language family, the improvement is very significant. Furthermore, the performance in the mTOP task is improved significantly by our method among all languages. This may be because that our method consider both syntax and lexicon simultaneously, which is beneficial for the semantic parsing task.

\subsection{Representation Similarities across Languages}
To evaluate the effectiveness of our method in aligning different languages, we employ the representation similarity between languages as the metric. Specifically, we utilize the testing set of XNLI \cite{conneau2018xnli} as the dataset, which consists of parallel sentences across multiple languages. Then we take the vector of [CLS] token from the final layer of our model, as well as the vectors from two baselines ("mBERT+Syn" and "mBERT+code-switch) for each sentence. Following \citet{libovicky2019language}, the centroid vector for representing each language is calculated by averaging these sentence representations. Finally, we adopt cosine similarity as the indicator to assess the degree of alignment between English and each target language.

\begin{figure}[htbp]
\centering
\includegraphics[width=0.48\textwidth,height=0.25\textwidth]{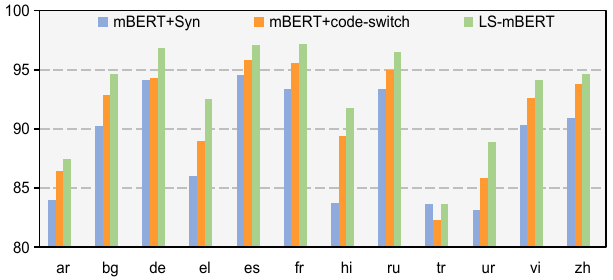} 
\caption{The similarities between languages. We first calculate the centroid representation for each language following \citet{libovicky2019language}. Then we adopt cosine similarity to evaluate the similarity between English and each target language. }
\label{fig.5}
\end{figure}

Figure~\ref{fig.5} illustrates the similarities between languages by using our method and the other two baselines. It can be easily found that our method outperforms the other two baselines in aligning language representations. This suggests that infusing two types of knowledge is indeed effective in reducing the disparities in language typologies, which improve cross-lingual transfer performance. In addition, we observe that "mBERT+code-switch" performs better than "mBERT+Syn", which reflects that lexical knowledge is more useful than syntactic knowledge for this task.

\subsection{Impact of Code-switching}
\begin{figure}[htbp]
\centering
\includegraphics[width=0.49\textwidth,height=0.26\textwidth]{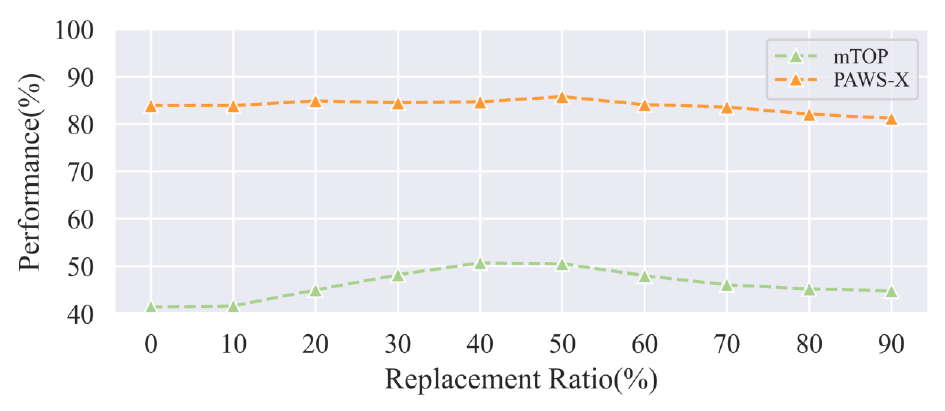} 
\caption{Performance on mTOP and PAWS-X with different replacement ratio $\alpha$ in code-switching.}
\label{fig.6}
\end{figure}

The replacement ratio $\alpha$ for code-switching is an important hyper-parameter in our method. Hence, we explore its impact on mTOP and PAWS-X, by varying $\alpha$ from 0 to 0.9 in increments of 0.1, shown in Figure ~\ref{fig.6}. When $\alpha$ is set to 0, it represents the results of the baseline "mBERT+Syn". As $\alpha$ increases, more source words are substituted with their equivalent words from other languages. The performance improvement certificates the effectiveness of code-switching technique. Notably, when about half of the words are replaced (0.5 for PAWS-X and 0.4 for mTOP), the performance reaches their peaks. After that, both tasks experience a decline in performance. This decline might be because the expression of meaning and sentence structure are influenced severely as too many words are replaced. Therefore, it is a optimal choice to set $\alpha$ between 0.4 to 0.5 for code-switching.

\begin{figure}[htbp]
\centering
\includegraphics[width=0.35\textwidth,height=0.27\textwidth]{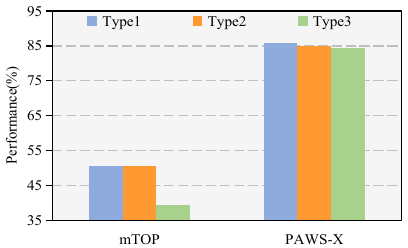} 
\caption{Performance on mTOP and PAWS-X with different replacement languages in code-switching. The source language for both tasks is English, and the results are averaged across all target languages excluding English. “Type1” represents the replacement with the target language; “Type2” represents the replacement with languages from the same language family as the target language; “Type3” represents the replacement with randomly selected languages.}
\label{fig.7}
\end{figure}

Furthermore, we investigate whether the choice of the replacement language in code-switching impacts our model's performance. We select mTOP and PAWS-X as the testing tasks. In code-switching, we devise three different measures for language replacement: "Exclusively replacing with the target language", "Replacing with languages from the same language family as the target language"; and "Replacing with languages selected randomly". The experimental results are illustrated in Figure ~\ref{fig.7}. We can easily observe that "Exclusively replacing with the target language" performs best, while "Replacing with randomly selected languages" yields the poorest results. Hence, this also underscores the importance of selecting languages closely related to each target language for substitution when employing the code-switching technique.

\begin{table*}[htbp]
\centering
\resizebox{\textwidth}{!}{
\begin{tabular}{llcccccccccccccccc}
\hline
\textbf{Task}&\textbf{Methods}& \multicolumn{1}{l}{\textbf{en}} & \multicolumn{1}{l}{\textbf{ar}} & \multicolumn{1}{l}{\textbf{bg}} & \multicolumn{1}{l}{\textbf{de}} & \multicolumn{1}{l}{\textbf{el}} & \multicolumn{1}{l}{\textbf{es}} & \multicolumn{1}{l}{\textbf{fr}} & \multicolumn{1}{l}{\textbf{hi}} & \multicolumn{1}{l}{\textbf{ru}} & \multicolumn{1}{l}{\textbf{tr}} & \multicolumn{1}{l}{\textbf{ur}} & \multicolumn{1}{l}{\textbf{vi}} & \multicolumn{1}{l}{\textbf{ko}} & \multicolumn{1}{l}{\textbf{nl}} & \multicolumn{1}{l}{\textbf{pt}} & \multicolumn{1}{l}{\textbf{AVG}} \\
\hline
\multirow{4}{*}{PAWS-X} & XLM-R& 84.2& 48.5& 80.5& 77.0 & 77.8 & 76.1 & 79.8 & 67.5 & 70.4 & 76.0 & 54.2 & 78.5  & 59.1 & 83.3 & 79.3 & 72.8\\
& XLM-R+Syn & 83.5 & 46.4 & 80.1 & 76.0 & 78.9 & 77.6 & 79.1 & 72.1 & 70.6 & 76.1 & 55.3 & 77.6 & 59.0 & 83.1 & 79.2 & 73.0 \\
& XKLM-R+code-switch & 83.4 & 46.8 & 81.7 & 78.2 & 79.2 & 71.1 & 78.6 & 72.9 & 70.6 & 77.2 & 57.9  & 76.0 & 58.2 & 83.6 & 80.0 & 73.0\\
& our method & 83.1 & 44.9 & 82.7 & 76.8 & 78.4 & 76.9 & 79.6 & 71.1 & 70.1 & 76.6 & 60.4 & 78.2 & 58.1 & 83.5 & 79.7  & 73.3\\
\hline
\end{tabular}
}
\caption{
Results for PAWS-X with XLM-R.
}
\label{table4}
\end{table*}

\subsection{Performance with XLM-R}
To validate the universality of our method, we substitute multilingual BERT with XLM-R in our framework. XLM-R is a more robust multilingual pre-trained model known for its  exceptional cross-lingual transfer capabilities. Subsequently, we test its performance on the PAWX-S dataset, and the experimental results are displayed in Table ~\ref{table4}.

In Table ~\ref{table4}, we also observe that our framework outperforms the other three baselines. This indicates that integrating lexical and syntactic knowledge is beneficial for enhancing performance, irrespective of the base model employed. Notably, our framework only achieves the slight performance improvement when utilizing XLM-R as the base model compared to employing multilingual BERT. It may be because that the base model, XLM-R, adopt larger corpus during pre-training, resulting in preserving richer language information. Consequently, XLM-R itself has possessed superior cross-lingual transfer capabilities. The assistance by incorporating external linguistic knowledge appears to be relatively minor in comparison.

\subsection{Limitations and Challenges}
In our study, we adopt a bilingual dictionary, such as MUSE \cite{lample2018word}, to substitute words in other languages. However, we randomly choose a target language word when there exist multiple translations for a source language word. This approach, although convenient, neglect the context of the source language word, potentially leading to inaccurate translations. This also highlights us to explore more precise word alignment methods in the future.

Furthermore, the tasks we have evaluated are quite limited, with some of them involving only a few languages. In the future, we will extend our method to more cross-lingual tasks. Meanwhile, we also develop dataset for these tasks to support more languages.

\section{Conclusion}
In this paper, we present a framework called "lexicon-syntax enhanced multilingual BERT" ("LS-mBERT"), which infuses lexical and syntactic knowledge to enhance cross-lingual transfer performance. Our method employs code-switching technology to generate input text mixed in various languages, enabling the entire model to capture lexical alignment information during training. Besides, a syntactic module consisting of a graph attention network (GAT) is introduced to guide mBERT in encoding language structures. The experimental results demonstrate that our proposed method outperforms all the baselines across different tasks, which certificates the effectiveness of integrating both types of knowledge into mBERT for improving cross-lingual transfer. In the future, we plan to incorporate different linguistic knowledge into large language models (LLMs) to further enhance cross-lingual transfer performance.

\section{Acknowledgements}
The authors would like to thank the anonymous reviewers for their feedback and suggestions. Additionally, this work was supported by the Major Program of the National Social Science Fund of China (18ZDA238), the National Social Science Fund of China (No.21CYY032), Beihang University Sponsored Projects for Core Young Researchers in the Disciplines of Social Sciences and Humanities(KG16183801) and Tianjin Postgraduate Scientific Research Innovation Program (No.2022BKY024).      


\nocite{*}
\section{Bibliographical References}\label{sec:reference}

\bibliographystyle{lrec-coling2024-natbib}
\bibliography{lrec-coling2024-example}

\begin{thebibliography}{41}
\expandafter\ifx\csname natexlab\endcsname\relax\def\natexlab#1{#1}\fi

\bibitem[{Ahmad et~al.(2021)Ahmad, Li, Chang, and Mehdad}]{ahmad2021syntax}
Wasi Ahmad, Haoran Li, Kai-Wei Chang, and Yashar Mehdad. 2021.
\newblock Syntax-augmented multilingual bert for cross-lingual transfer.
\newblock In \emph{Proceedings of the 59th Annual Meeting of the Association
  for Computational Linguistics and the 11th International Joint Conference on
  Natural Language Processing (Volume 1: Long Papers)}, pages 4538--4554.

\bibitem[{Asai et~al.(2021)Asai, Kasai, Clark, Lee, Choi, and
  Hajishirzi}]{asai2021xor}
Akari Asai, Jungo Kasai, Jonathan~H Clark, Kenton Lee, Eunsol Choi, and
  Hannaneh Hajishirzi. 2021.
\newblock Xor qa: Cross-lingual open-retrieval question answering.
\newblock In \emph{Proceedings of the 2021 Conference of the North American
  Chapter of the Association for Computational Linguistics: Human Language
  Technologies}, pages 547--564.

\bibitem[{Chen et~al.(2022)Chen, Shou, Gong, and Pei}]{chen2022good}
Nuo Chen, Linjun Shou, Ming Gong, and Jian Pei. 2022.
\newblock From good to best: Two-stage training for cross-lingual machine
  reading comprehension.
\newblock In \emph{Proceedings of the AAAI Conference on Artificial
  Intelligence}, volume~36, pages 10501--10508.

\bibitem[{Cignarella et~al.(2020)Cignarella, Basile, Sanguinetti, Bosco, Rosso,
  and Benamara}]{cignarella2020multilingual}
Alessandra~Teresa Cignarella, Valerio Basile, Manuela Sanguinetti, Cristina
  Bosco, Paolo Rosso, and Farah Benamara. 2020.
\newblock Multilingual irony detection with dependency syntax and neural
  models.
\newblock In \emph{Proceedings of the 28th International Conference on
  Computational Linguistics}, pages 1346--1358.

\bibitem[{Conneau et~al.(2019)Conneau, Khandelwal, Goyal, Chaudhary, Wenzek,
  Guzm{\'a}n, Grave, Ott, Zettlemoyer, and Stoyanov}]{conneau2019unsupervised}
Alexis Conneau, Kartikay Khandelwal, Naman Goyal, Vishrav Chaudhary, Guillaume
  Wenzek, Francisco Guzm{\'a}n, Edouard Grave, Myle Ott, Luke Zettlemoyer, and
  Veselin Stoyanov. 2019.
\newblock Unsupervised cross-lingual representation learning at scale.
\newblock \emph{arXiv preprint arXiv:1911.02116}.

\bibitem[{Conneau et~al.(2018)Conneau, Rinott, Lample, Williams, Bowman,
  Schwenk, and Stoyanov}]{conneau2018xnli}
Alexis Conneau, Ruty Rinott, Guillaume Lample, Adina Williams, Samuel Bowman,
  Holger Schwenk, and Veselin Stoyanov. 2018.
\newblock Xnli: Evaluating cross-lingual sentence representations.
\newblock In \emph{Proceedings of the 2018 Conference on Empirical Methods in
  Natural Language Processing}, pages 2475--2485.

\bibitem[{de~Vries et~al.(2022)de~Vries, Wieling, and Nissim}]{de2022make}
Wietse de~Vries, Martijn Wieling, and Malvina Nissim. 2022.
\newblock Make the best of cross-lingual transfer: Evidence from pos tagging
  with over 100 languages.
\newblock In \emph{Proceedings of the 60th Annual Meeting of the Association
  for Computational Linguistics (Volume 1: Long Papers)}, pages 7676--7685.

\bibitem[{Ellis and Larsen-Freeman(2009)}]{ellis2009language}
Nick~C Ellis and Diane Larsen-Freeman. 2009.
\newblock \emph{Language as a complex adaptive system}, volume~11.
\newblock John Wiley \& Sons.

\bibitem[{Fetahu et~al.(2022)Fetahu, Fang, Rokhlenko, and
  Malmasi}]{fetahu2022dynamic}
Besnik Fetahu, Anjie Fang, Oleg Rokhlenko, and Shervin Malmasi. 2022.
\newblock Dynamic gazetteer integration in multilingual models for
  cross-lingual and cross-domain named entity recognition.
\newblock In \emph{Proceedings of the 2022 Conference of the North American
  Chapter of the Association for Computational Linguistics: Human Language
  Technologies}, pages 2777--2790.

\bibitem[{Han et~al.(2022)Han, Luo, Chen, Liu, Sun, Botong, Fei, and
  Zheng}]{han2022cross}
Xu~Han, Yuqi Luo, Weize Chen, Zhiyuan Liu, Maosong Sun, Zhou Botong, Hao Fei,
  and Suncong Zheng. 2022.
\newblock Cross-lingual contrastive learning for fine-grained entity typing for
  low-resource languages.
\newblock In \emph{Proceedings of the 60th Annual Meeting of the Association
  for Computational Linguistics (Volume 1: Long Papers)}, pages 2241--2250.

\bibitem[{He et~al.(2019)He, Li, and Zhao}]{he2019syntax}
Shexia He, Zuchao Li, and Hai Zhao. 2019.
\newblock Syntax-aware multilingual semantic role labeling.
\newblock In \emph{Proceedings of the 2019 Conference on Empirical Methods in
  Natural Language Processing and the 9th International Joint Conference on
  Natural Language Processing (EMNLP-IJCNLP)}, pages 5350--5359.

\bibitem[{Hsu et~al.(2019)Hsu, Liu, and Lee}]{hsu2019zero}
Tsung-Yuan Hsu, Chi-Liang Liu, and Hung-Yi Lee. 2019.
\newblock Zero-shot reading comprehension by cross-lingual transfer learning
  with multi-lingual language representation model.
\newblock In \emph{Proceedings of the 2019 Conference on Empirical Methods in
  Natural Language Processing and the 9th International Joint Conference on
  Natural Language Processing (EMNLP-IJCNLP)}, pages 5933--5940.

\bibitem[{Hu et~al.(2020)Hu, Ruder, Siddhant, Neubig, Firat, and
  Johnson}]{hu2020xtreme}
Junjie Hu, Sebastian Ruder, Aditya Siddhant, Graham Neubig, Orhan Firat, and
  Melvin Johnson. 2020.
\newblock Xtreme: A massively multilingual multi-task benchmark for evaluating
  cross-lingual generalisation.
\newblock In \emph{International Conference on Machine Learning}, pages
  4411--4421. PMLR.

\bibitem[{Kenton and Toutanova(2019)}]{kenton2019bert}
Jacob Devlin Ming-Wei~Chang Kenton and Lee~Kristina Toutanova. 2019.
\newblock Bert: Pre-training of deep bidirectional transformers for language
  understanding.
\newblock In \emph{Proceedings of naacL-HLT}, volume~1, page~2.

\bibitem[{Kim et~al.(2017)Kim, Kim, Sarikaya, and
  Fosler-Lussier}]{kim2017cross}
Joo-Kyung Kim, Young-Bum Kim, Ruhi Sarikaya, and Eric Fosler-Lussier. 2017.
\newblock Cross-lingual transfer learning for pos tagging without cross-lingual
  resources.
\newblock In \emph{Proceedings of the 2017 conference on empirical methods in
  natural language processing}, pages 2832--2838.

\bibitem[{Lai et~al.(2021)Lai, Huang, Jing, Chen, Xu, and
  Liu}]{lai2021saliency}
Siyu Lai, Hui Huang, Dong Jing, Yufeng Chen, Jinan Xu, and Jian Liu. 2021.
\newblock Saliency-based multi-view mixed language training for zero-shot
  cross-lingual classification.
\newblock In \emph{Findings of the Association for Computational Linguistics:
  EMNLP 2021}, pages 599--610.

\bibitem[{Lample et~al.(2018)Lample, Conneau, Ranzato, Denoyer, and
  J{\'e}gou}]{lample2018word}
Guillaume Lample, Alexis Conneau, Marc'Aurelio Ranzato, Ludovic Denoyer, and
  Herv{\'e} J{\'e}gou. 2018.
\newblock Word translation without parallel data.
\newblock In \emph{International Conference on Learning Representations}.

\bibitem[{Langedijk et~al.(2022)Langedijk, Dankers, Lippe, Bos, Guevara,
  Yannakoudakis, and Shutova}]{langedijk2022meta}
Anna Langedijk, Verna Dankers, Phillip Lippe, Sander Bos, Bryan~Cardenas
  Guevara, Helen Yannakoudakis, and Ekaterina Shutova. 2022.
\newblock Meta-learning for fast cross-lingual adaptation in dependency
  parsing.
\newblock In \emph{Proceedings of the 60th Annual Meeting of the Association
  for Computational Linguistics (Volume 1: Long Papers)}, pages 8503--8520.

\bibitem[{Li et~al.(2021)Li, Arora, Chen, Gupta, Gupta, and
  Mehdad}]{li2021mtop}
Haoran Li, Abhinav Arora, Shuohui Chen, Anchit Gupta, Sonal Gupta, and Yashar
  Mehdad. 2021.
\newblock Mtop: A comprehensive multilingual task-oriented semantic parsing
  benchmark.
\newblock In \emph{Proceedings of the 16th Conference of the European Chapter
  of the Association for Computational Linguistics: Main Volume}, pages
  2950--2962.

\bibitem[{Li et~al.(2023)Li, Zhang, Wang, Li, Jatowt, and Yang}]{li2023across}
Peiyao Li, Zhengkun Zhang, Jun Wang, Liang Li, Adam Jatowt, and Zhenglu Yang.
  2023.
\newblock Across: An alignment-based framework for low-resource many-to-one
  cross-lingual summarization.
\newblock In \emph{Findings of the Association for Computational Linguistics:
  ACL 2023}, pages 2458--2472.

\bibitem[{Liang et~al.(2022)Liang, Meng, Zhou, Xu, Chen, Su, and
  Zhou}]{liang2022variational}
Yunlong Liang, Fandong Meng, Chulun Zhou, Jinan Xu, Yufeng Chen, Jinsong Su,
  and Jie Zhou. 2022.
\newblock A variational hierarchical model for neural cross-lingual
  summarization.
\newblock In \emph{Proceedings of the 60th Annual Meeting of the Association
  for Computational Linguistics (Volume 1: Long Papers)}, pages 2088--2099.

\bibitem[{Libovick{\`y} et~al.(2019)Libovick{\`y}, Rosa, and
  Fraser}]{libovicky2019language}
Jind{\v{r}}ich Libovick{\`y}, Rudolf Rosa, and Alexander Fraser. 2019.
\newblock How language-neutral is multilingual bert?
\newblock \emph{arXiv preprint arXiv:1911.03310}.

\bibitem[{Nie et~al.(2022)Nie, Liang, Schmid, and Sch{\"u}tze}]{nie2022cross}
Ercong Nie, Sheng Liang, Helmut Schmid, and Hinrich Sch{\"u}tze. 2022.
\newblock Cross-lingual retrieval augmented prompt for low-resource languages.
\newblock \emph{arXiv preprint arXiv:2212.09651}.

\bibitem[{Nooralahzadeh and Sennrich(2023)}]{nooralahzadeh2023improving}
Farhad Nooralahzadeh and Rico Sennrich. 2023.
\newblock Improving the cross-lingual generalisation in visual question
  answering.
\newblock In \emph{Proceedings of the AAAI Conference on Artificial
  Intelligence}, volume~37, pages 13419--13427.

\bibitem[{Pan et~al.(2017)Pan, Zhang, May, Nothman, Knight, and
  Ji}]{pan2017cross}
Xiaoman Pan, Boliang Zhang, Jonathan May, Joel Nothman, Kevin Knight, and Heng
  Ji. 2017.
\newblock Cross-lingual name tagging and linking for 282 languages.
\newblock In \emph{Proceedings of the 55th Annual Meeting of the Association
  for Computational Linguistics (Volume 1: Long Papers)}, pages 1946--1958.

\bibitem[{Qi et~al.(2022)Qi, Wan, Du, and Chen}]{qi2022enhancing}
Kunxun Qi, Hai Wan, Jianfeng Du, and Haolan Chen. 2022.
\newblock Enhancing cross-lingual natural language inference by prompt-learning
  from cross-lingual templates.
\newblock In \emph{Proceedings of the 60th Annual Meeting of the Association
  for Computational Linguistics (Volume 1: Long Papers)}, pages 1910--1923.

\bibitem[{Qi et~al.(2020)Qi, Zhang, Zhang, Bolton, and Manning}]{qi2020stanza}
Peng Qi, Yuhao Zhang, Yuhui Zhang, Jason Bolton, and Christopher~D Manning.
  2020.
\newblock Stanza: A python natural language processing toolkit for many human
  languages.
\newblock \emph{arXiv preprint arXiv:2003.07082}.

\bibitem[{Qin et~al.(2021)Qin, Ni, Zhang, and Che}]{qin2021cosda}
Libo Qin, Minheng Ni, Yue Zhang, and Wanxiang Che. 2021.
\newblock Cosda-ml: multi-lingual code-switching data augmentation for
  zero-shot cross-lingual nlp.
\newblock In \emph{Proceedings of the Twenty-Ninth International Conference on
  International Joint Conferences on Artificial Intelligence}, pages
  3853--3860.

\bibitem[{Shi et~al.(2022)Shi, Gimpel, and Livescu}]{shi2022substructure}
Freda Shi, Kevin Gimpel, and Karen Livescu. 2022.
\newblock Substructure distribution projection for zero-shot cross-lingual
  dependency parsing.
\newblock In \emph{Proceedings of the 60th Annual Meeting of the Association
  for Computational Linguistics (Volume 1: Long Papers)}, pages 6547--6563.

\bibitem[{Straka and Strakov{\'a}(2017)}]{straka2017tokenizing}
Milan Straka and Jana Strakov{\'a}. 2017.
\newblock Tokenizing, pos tagging, lemmatizing and parsing ud 2.0 with udpipe.
\newblock In \emph{Proceedings of the CoNLL 2017 shared task: Multilingual
  parsing from raw text to universal dependencies}, pages 88--99.

\bibitem[{Veli{\v{c}}kovi{\'c} et~al.(2017)Veli{\v{c}}kovi{\'c}, Cucurull,
  Casanova, Romero, Lio, and Bengio}]{velivckovic2017graph}
Petar Veli{\v{c}}kovi{\'c}, Guillem Cucurull, Arantxa Casanova, Adriana Romero,
  Pietro Lio, and Yoshua Bengio. 2017.
\newblock Graph attention networks.
\newblock \emph{arXiv preprint arXiv:1710.10903}.

\bibitem[{Wang and Zheng(2015)}]{wang2015transfer}
Dong Wang and Thomas~Fang Zheng. 2015.
\newblock Transfer learning for speech and language processing.
\newblock In \emph{2015 Asia-Pacific Signal and Information Processing
  Association Annual Summit and Conference (APSIPA)}, pages 1225--1237. IEEE.

\bibitem[{Wang et~al.(2022)Wang, Ruder, and Neubig}]{wang2022expanding}
Xinyi Wang, Sebastian Ruder, and Graham Neubig. 2022.
\newblock Expanding pretrained models to thousands more languages via
  lexicon-based adaptation.
\newblock In \emph{Proceedings of the 60th Annual Meeting of the Association
  for Computational Linguistics (Volume 1: Long Papers)}.

\bibitem[{Wang et~al.(2021)Wang, Liu, Yang, Liu, and Wang}]{wang2021cross}
Ziyun Wang, Xuan Liu, Peiji Yang, Shixing Liu, and Zhisheng Wang. 2021.
\newblock Cross-lingual text classification with heterogeneous graph neural
  network.
\newblock In \emph{Proceedings of the 59th Annual Meeting of the Association
  for Computational Linguistics and the 11th International Joint Conference on
  Natural Language Processing (Volume 2: Short Papers)}, pages 612--620.

\bibitem[{Wu et~al.(2022)Wu, Wu, Zhang, Xiong, Chen, Zhuang, and
  Feng}]{wu2022learning}
Linjuan Wu, Shaojuan Wu, Xiaowang Zhang, Deyi Xiong, Shizhan Chen, Zhiqiang
  Zhuang, and Zhiyong Feng. 2022.
\newblock Learning disentangled semantic representations for zero-shot
  cross-lingual transfer in multilingual machine reading comprehension.
\newblock In \emph{Proceedings of the 60th Annual Meeting of the Association
  for Computational Linguistics (Volume 1: Long Papers)}, pages 991--1000.

\bibitem[{Xie et~al.(2018)Xie, Yang, Neubig, Smith, and
  Carbonell}]{xie2018neural}
Jiateng Xie, Zhilin Yang, Graham Neubig, Noah~A Smith, and Jaime~G Carbonell.
  2018.
\newblock Neural cross-lingual named entity recognition with minimal resources.
\newblock In \emph{Proceedings of the 2018 Conference on Empirical Methods in
  Natural Language Processing}, pages 369--379.

\bibitem[{Xu et~al.(2022)Xu, Zhang, Zong, Liu, Cheng, Ni, Chen, Zhao, and
  Choi}]{xu2022zero}
Liyan Xu, Xuchao Zhang, Bo~Zong, Yanchi Liu, Wei Cheng, Jingchao Ni, Haifeng
  Chen, Liang Zhao, and Jinho~D Choi. 2022.
\newblock Zero-shot cross-lingual machine reading comprehension via
  inter-sentence dependency graph.
\newblock In \emph{Proceedings of the AAAI Conference on Artificial
  Intelligence}, volume~36, pages 11538--11546.

\bibitem[{Yang et~al.(2019)Yang, Zhang, Tar, and Baldridge}]{yang2019paws}
Yinfei Yang, Yuan Zhang, Chris Tar, and Jason Baldridge. 2019.
\newblock Paws-x: A cross-lingual adversarial dataset for paraphrase
  identification.
\newblock \emph{arXiv preprint arXiv:1908.11828}.

\bibitem[{Yu et~al.(2021)Yu, Zhang, Niu, Sun, and Jiang}]{yu2021cosy}
Sicheng Yu, Hao Zhang, Yulei Niu, Qianru Sun, and Jing Jiang. 2021.
\newblock Cosy: Counterfactual syntax for cross-lingual understanding.
\newblock In \emph{Proceedings of the 59th Annual Meeting of the Association
  for Computational Linguistics and the 11th International Joint Conference on
  Natural Language Processing (Volume 1: Long Papers)}, pages 577--589.

\bibitem[{Zhang et~al.(2021{\natexlab{a}})Zhang, Zhang, Ye, Li, Sun, Zhu, Zhao,
  and Zhang}]{zhang2021point}
Tong Zhang, Long Zhang, Wei Ye, Bo~Li, Jinan Sun, Xiaoyu Zhu, Wen Zhao, and
  Shikun Zhang. 2021{\natexlab{a}}.
\newblock Point, disambiguate and copy: Incorporating bilingual dictionaries
  for neural machine translation.
\newblock In \emph{Proceedings of the 59th Annual Meeting of the Association
  for Computational Linguistics and the 11th International Joint Conference on
  Natural Language Processing (Volume 1: Long Papers)}, pages 3970--3979.

\bibitem[{Zhang et~al.(2021{\natexlab{b}})Zhang, Strubell, and
  Hovy}]{zhang2021benefit}
Zhisong Zhang, Emma Strubell, and Eduard Hovy. 2021{\natexlab{b}}.
\newblock On the benefit of syntactic supervision for cross-lingual transfer in
  semantic role labeling.
\newblock In \emph{Proceedings of the 2021 Conference on Empirical Methods in
  Natural Language Processing}, pages 6229--6246.

\end{thebibliography}


\end{document}